%% file: main.tex
\pdfoutput=1

\documentclass[11pt]{article}

\usepackage{emnlp2021}

\usepackage{times}
\usepackage{latexsym}
\usepackage{pifont}
\newcommand{\cmark}{\ding{51}}%
\newcommand{\xmark}{\ding{55}}%

\usepackage[T1]{fontenc}

\usepackage[utf8]{inputenc}

\usepackage{microtype}

\input{math_commands.tex}

\usepackage{hyperref}
\usepackage{url}

\usepackage{amsmath,amssymb,amsthm}
\usepackage[utf8]{inputenc}
\usepackage[abbreviations,british]{foreign}
\usepackage[colorinlistoftodos,textsize=tiny]{todonotes}
\usepackage{booktabs}
\usepackage{multirow,hhline,graphicx,array,tabulary}

\graphicspath{ {./img/} }

\title{End-to-End Entity Resolution and Question Answering Using Differentiable Knowledge Graphs}

\author{Armin Oliya \qquad Amir Saffari \qquad Priyanka Sen \qquad Tom Ayoola \\
Amazon Alexa AI \\ Cambridge, UK \\
\texttt{\{aooliya,amsafari,sepriyan,tayoola\}@amazon.com}
}

\begin{document}
\maketitle
\begin{abstract}
Recently, end-to-end (E2E) trained models for question answering over knowledge graphs (KGQA) have delivered promising results using only a weakly supervised dataset. However, these models are trained and evaluated in a setting where hand-annotated question entities are supplied to the model, leaving the important and non-trivial task of entity resolution (ER) outside the scope of E2E learning. In this work, we extend the boundaries of E2E learning for KGQA to include the training of an ER component. Our model only needs the question text and the answer entities to train, and delivers a stand-alone QA model that does not require an additional ER component to be supplied during runtime. Our approach is fully differentiable, thanks to its reliance on a recent method for building differentiable KGs \citep{Cohen2020ICLR}. We evaluate our E2E trained model on two public datasets and show that it comes close to baseline models that use hand-annotated entities.



\end{abstract}

\section{Introduction}



The conventional approach for Question Answering using a Knowledge Graph (KGQA) involves a set of loosely connected components; notably, an entity resolution component identifies entities mentioned in the question, and a semantic parsing component produces a structured representation of the question. The programs resulting from combining these components can be executed on a knowledge graph (KG) engine to retrieve the answers.


While this approach can be effective, collecting training datasets for individual components can be challenging \cite{dahl1994, finegan-dollak-etal-2018-improving}. For example, supervised semantic parsing requires training data pairing natural-language questions with structured queries, which is difficult to obtain. This has motivated many efforts in weakly supervised training \cite{https://doi.org/10.1002/widm.1389}. Following recent breakthroughs in machine translation \cite{DBLP:journals/corr/BahdanauCB14}, a new goal is to directly optimize the entire chain of components end-to-end, without the need for intermediate annotations. 




\begin{figure}[t]

\includegraphics[width=\columnwidth]{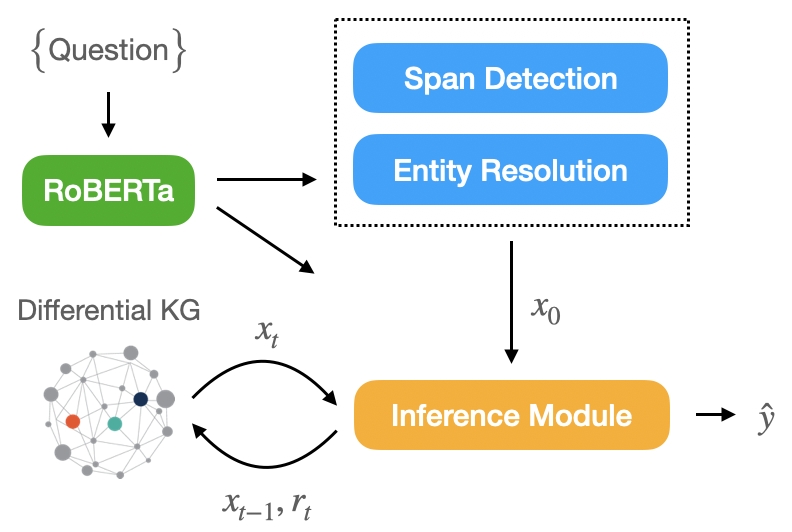}
\caption{High-level architecture of the end-to-end model. One forward pass of RoBERTa extracts contextual embeddings for all components. Span Detection and Entity Resolution happen jointly to derive seed entities vector $\bm{x_0}$. The inference module performs multi-hop reasoning to reach answer entities vector $\hat{\bm{y}}$ } 
\label{fig:highlevel}

\end{figure}

However, entity resolution (ER) is by and large a neglected component of E2E learning, and existing weakly supervised solutions mostly assume question entities are either given or extracted by an external system. In practice, there's a scarcity of quality training data for ER on questions, and poor entity extraction by out-of-domain models affects the overall performance of a KGQA system \citep{SINGH2020100594,han-etal-2020-empirical}.

In this work, we present an end-to-end model for KGQA that learns to jointly perform entity resolution and inference. Our work leverages the differentiable KG proposed in \newcite{Cohen2020ICLR}, which allows all the components of our model to be trained using a dataset of only questions and answers. This eliminates the need for labelled ER data for questions and allows our model to run independently, without relying on external components. Furthermore, the tight integration of ER into our solution allows uncertainties about entities to be directly reflected in our confidence in answers. 

\section{Related Work}

Traditional approaches to KGQA rely on semantic parsing (SP) to translate natural language into a logical form. Weakly supervised SP is a well-studied topic with increasing interest in applying Reinforcement Learning (RL) \cite{HUA2020100612,pmlr-v97-agarwal19e}. ER is rarely considered in the scope of surveyed solutions, and if it is, it's treated as an independent component and not included in the weak supervision scope \cite{ijcai2019-679}. In general, RL algorithms for QA are hard to tune and have large variances in their results. The exploration-exploitation issue also lets models settle on high‐reward but spurious logical forms, leading to poor generalization \cite{https://doi.org/10.1002/widm.1389}. Including ER with a discrete output space as part of an E2E RL pipeline will further add to the challenges that RL-based solutions face.


Another line of work in KGQA uses embedding techniques to implicitly infer answers from knowledge graphs without explicit queries \cite{saxena-etal-2020-improving, sun-etal-2019-pullnet}. While these embedding-based approaches perform well, they are memory intensive and difficult to scale to large knowledge graphs. In addition, when new entities are added to the KG, they need to be retrained to learn updated embeddings. The differentiable KG we use in this work can incorporate new entities without affecting trained models, and can scale to billions of entities via horizontal scaling \citep{Cohen2020ICLR}.



The few relevant works on entity resolution for questions utilize complex models with many interworking modules (e.g. \citealt{sorokin-gurevych-2018-mixing, tan-etal-2017-entity}). ELQ \citep{li-etal-2020-efficient} is a more recent effort and simplifies the process by relying on a bi-encoder to jointly perform span detection and ER in a multi-task setup. However, these solutions rely on direct supervision. Our proposed method eliminates the need for labelled data for ER. In fact, the weakly supervised ER model presented here could be detached and used as a standalone ER module after training.

\section{E2E Model}\label{sec:problemdef}




\subsection{Differentiable Knowledge Graph}\label{sec:nkg}

A traditional knowledge graph (KG) stores facts as triples and uses a symbolic query engine to extract answers. A differentiable KG stores facts in tensors and makes query execution over facts differentiable. 



We use the approach presented in ReifiedKB \cite{Cohen2020ICLR} to create a scalable and differentiable knowledge graph supporting multi-hop relation following programs. We provide an overview here but full details can be found in the original paper. Assume the set of all triples in a knowledge graph $\mathcal{T} = \{t_i\}_{i=1}^{N_T}, \quad t_i = (s_{s_i}, p_{p_i}, o_{o_i})$ are represented by the following sparse matrices:
\begin{align*}
  M_s \in \{0, 1\}^{N_T \times N_E}, & \quad M_s(i, j) = \mathbb{I} \bigl( e_j = s_{s_i} \bigr) \\
  M_o \in \{0, 1\}^{N_T \times N_E}, & \quad M_o(i, j) = \mathbb{I} \bigl( e_j = o_{o_i} \bigr) \\
  M_p \in \{0, 1\}^{N_T \times N_R}, & \quad M_p(i, j) = \mathbb{I} \bigl( p_j = p_{p_i} \bigr), 
\end{align*}

\noindent where $M_s$, $M_o$, and $M_p$ are denoted as subject, object, and relation index matrices. $N_T$, $N_E$, and $N_R$ are the number of triples, entities, and relations, respectively. 

Given an entities vector $\mathbf{x}_{t - 1} \in \mathbb{R}^{N_E} $ at $t - 1$-th hop, the entities vector $\mathbf{x}_t$ resulting from following
a relation vector $\mathbf{r}_{t}\in \mathbb{R}^{N_R}$ can be computed by:
\begin{equation}\label{eq:reified-follow-kgqa}
  \mathbf{x}_t = \text{follow}(\mathbf{x}_{t - 1}, \mathbf{r}_t) = M_o^T ( M_s \mathbf{x}_{t - 1} \odot  M_p \mathbf{r}_t ) ,
\end{equation}

\noindent where $\odot$ is the element-wise multiplication. 

\subsection{Multi-hop Inference}\label{sec:inference}

Given an input question $q = q_1 \cdots  q_n$ of length $n$, we first use the pretrained language model RoBERTa \cite{Liu2019RoBERTaAR} to extract contextual embeddings for each token:
\begin{equation}\label{eq:cx-embedding}
  [\mathbf{h_{q}}, \bm{q_{1}} \cdots  \bm{q_{n}}]^T = LM(CLS, q_1 \cdots q_n) 
\end{equation}

\noindent where $\bm{h_{q}} \in \mathbb{R}^{D}$ corresponds to the $CLS$ token and is used as the question embedding. We  compute the relation vector for the $t$-th hop using a hierarchical decoder and subsequent entities vector by following that vector as:
\begin{align}\label{eq:rigel-inference}
  \mathbf{r}_t = & \text{softmax} \Bigl( \mathbf{W_t^{inf}} \bigl[ \mathbf{h}_q | \mathbf{r}_{t - 1} | \cdots | \mathbf{r}_{1} \bigr]^T \Bigr) ,\\
  \label{eq:rigel-inference-2}
  \mathbf{x}_t = & \text{follow}(\mathbf{x}_{t - 1}, \mathbf{r}_t) .
\end{align}

Since the follow operation (Equation \ref{eq:reified-follow-kgqa}) can be chained infinitely, we set a maximum number of hops and use an attention mechanism to combine answer entities across all hops. We compute attention across all hops by:
\begin{align}\label{eq:rigel-attention}
  \mathbf{c_t} = & \mathbf{W_t^{att}} \bigl[ \mathbf{h}_q | \mathbf{r}_{t - 1} | \cdots | \mathbf{r}_{1} \bigr]^T , \\
  \mathbf{a} = & \text{softmax} ([ \mathbf{c_1}, \cdots,  \mathbf{c_{T_{max}}} ]) ,
\end{align}
where $T_{max}$ is the predefined maximum number of hops. The final answer entities vector will be:
\begin{equation}\label{eq:rigel-answer}
  \hat{\mathbf{y}} = \sum_{t = 1}^{T_{max}} a_t \mathbf{x}_t .
\end{equation}

Compared to ReifiedKB, our decoder uses RoBERTa for embedding questions, simplifies the stopping mechanism, and allows returning more than one answer entity. 

\subsection{Entity Resolution}\label{sec:er}



We approach Entity Resolution by estimating the likelihood of all the plausible spans in the question and then selecting the most likely candidate entity for each span.

Given an input question $q = q_1 \cdots  q_n$ of length $n$, the likelihood of each span $[i, j]$ in the question ($i$-th to $j$-th tokens of $q$) is calculated as:
\begin{align}\label{eq:md}
  \bm{q_{ij}} = & \frac{\sum_i^j \bm{q_k}}{(j-i+1)} \in \mathbb{R}^{D} , \\
  \label{eq:md2}
  s_{ij} = & \frac{\exp(\bm{q_{ij}}  \bm{w_{s}})}{\sum_{\forall m,n} \exp(\bm{q_{mn}}\bm{w_{s}})}  ,
\end{align}

\noindent where $\bm{w_{s}} \in \mathbb{R}^{D \times 1}$ is a learnable matrix and $\bm{q_k}$ are contextual token embeddings from Equation \ref{eq:cx-embedding}.

For a given span, candidate entities that could be referred to by that span are extracted by exact search against a lookup table, built using titles and aliases of entities in the KG. Candidate generation can further be improved by considering other approximate or fuzzy search methods, but we leave this as future work. 

If there are overlapping candidates between two spans, they are assigned to the longer one. For example, consider the question "what position does carlos gomez play?". If candidates of the three spans "carlos", "gomez", and "carlos gomez" all contain \code{Q2747238} (the Wikidata entity ID referring to Carlos Gomez, the Dominican baseball player), the entity ID will be assigned to the longest span only ("carlos gomez"). This is to avoid having duplicate entities across spans and comes from the intuition that longer spans are more specific and should be preferred. We have not seen errors arising from this preprocessing step. 

Assume ${c}_{ij}^k$ is the $k$-th candidate for span $[i,j]$. We embed each candidate entity by learning a dense representation of its KG neighbourhood:
\begin{align}\label{eq:er-cand-encoding}
  \mathcal{F}_{ij}^k = & \{f_{em}([p|o]) \; \forall \; ({c}_{ij}^k, p, o) \in \mathcal{G}\} ,\\
  \mathbf{\mathbf{z_{ij}^k}} = & \frac{ \sum_{\bm{\upsilon} \in \mathcal{F}_{ij}^k} \bm{\upsilon}} {|\mathcal{F}_{ij}^k|} \; \in \mathbb{R}^{D} ,
\end{align}



\noindent where $\mathcal{G}$ is the knowledge graph, $[p|o]$ is the string concatenation of $p$ and $o$, and $f_{em}$ is an embedding function that maps a string to a dense vector. 

For example, assume \code{Q2747238} is the candidate entity we want to embed. It is on the left-hand side of two triples in the knowledge graph: \code{(Q2747238, instance-of, human)} and \code{(Q2747238, occupation, baseball player)}. 

We first create the two strings \textit{“instance-of$\colon$human''} and \textit{“occupation$\colon$baseball player''} and pass them to $f_{em}$ to embed. These strings are treated as features of \code{Q2747238}, and we are able to learn embeddings effectively since they are also used for other humans and baseball players. Finally, we take the average of these two feature embeddings to get to the entity embedding for \code{Q2747238}. These operations are implemented using \code{torch.nn.EmbeddingBag} in PyTorch with random initialization. Our approach is not limited by knowledge graph features or this specific embedding approach; for instance, a RoBERTa encoding of entity descriptions could be used as an alternative. We leave experiments with other entity representations as future work.


Given the embedding, the likelihood of a candidate entity is estimated by considering the span likelihood and the likelihood of other candidates in that span:




\begin{equation}\label{eq:er-score}
  e_{ij}^k = s_{ij} \frac{\exp([\mathbf{z_{ij}^k}  \centerdot \bm{q_{ij}} ])}{\sum_{\forall l} \exp([\mathbf{z_{ij}^l}  \centerdot \bm{q_{ij}} ])} .
\end{equation}




To get to the final entity vector, we re-score candidate entities across all spans:
\begin{align}\label{eq:rigeler-final-score}
  x_{ij}^k = & \frac{\exp(e_{ij}^k)}{\sum_{\forall u,v,w}  \exp(e_{uv}^w)} , \\
  \mathbf{x_{0}} = & x_{ij}^k  \mapsto \vec{0}  \in \mathbb{R}^{N_E}, 
\end{align}


\noindent where $\mapsto$ maps each candidate entity likelihood to its corresponding index in the zero vector $\vec{0}$. The resulting $\mathbf{x}_0$ vector is used in Equation \ref{eq:rigel-inference-2} and captures uncertainties about entity resolution. It  is different from \citealt{Cohen2020ICLR} where $\mathbf{x}_0$ is assumed to be given with $\{0,1\}$ as the only possible values.

\subsection{Training}\label{sec:training}

We train the model using the binary cross-entropy loss function:
\begin{equation}\label{eq:rigel-loss}
  \mathcal{L}(\mathbf{y}, \mathbf{\hat{y}}) = \frac{1}{N_E} \sum_{i = 1}^{N_E} y_i \log \hat{y}_i + (1 - y_i) \log (1 - \hat{y}_i) ,
\end{equation}

\noindent where ${\mathbf{y}}\in\mathbb{R}^{N_E}$ is a \textit{k}-hot label vector. While ReifiedKB uses cross-entropy loss, we instead use a multi-label loss function across all entities. This is because the output space in a majority of cases contains multiple entities, so cross-entropy loss is inadequate. During training, the entity resolution and inference modules are trained jointly and uncertainties about each module are propagated to final answer entities vector ${\mathbf{\hat{y}}}$.





\section{Experiments}\label{sec:experiments}
We call our model \textit{Rigel} and evaluate 3 versions at different E2E learning boundaries. The baseline model , \textit{Rigel-Baseline}, is given gold entities and no entity resolution is involved, demonstrating the performance of the inference module alone. \textit{Rigel-ER} is given the gold spans, but still has to learn to disambiguate between candidate entities for that span. Finally, in \textit{Rigel-E2E}, we provide the question text only, requiring the model to attend to the right span \textit{and} disambiguate between candidates for each span.

\subsection{Datasets}
We evaluate our models on two open-domain Question Answering datasets: SimpleQuestions \citep{DBLP:journals/corr/BordesUCW15} and WebQSP  \citep{yih-etal-2016-value}. Both datasets were constructed based on the outdated FreeBase. Therefore, to generate better candidates and entity representation, we chose to use a subset of these datasets that are answerable by Wikidata \cite{wikidata-benchmark}. This is different from other baselines we compare against, which do not include an ER component. For WebQSP, this leads to $2349$ train, $261$ dev, and $1375$ test set samples. For SimpleQuestions the number of samples are $19471$ train, $2818$ dev, and $5620$ test.

Questions in SimpleQuestions and WebQSP can be answered in 1 and 2 hops respectively, so we set the maximum number of hops $T_{max}$ in Equation \ref{eq:rigel-answer} accordingly. For each dataset, we also limit Wikidata to a subset that is $T_{max}$-hop reachable from any of the candidates $c_{ij}^k$ in Equation \ref{eq:er-cand-encoding}. This results in a subgraph with 3.7 million triples, 1.0 million entities, and 1,158 relations for SimpleQuestions; and 4.9 million triples, 1.1 million entities, and 1,230 relations for WebQSP.






\subsection{Results}\label{sec:rigel-experiments}

Results of our experiments are shown in Table \ref{tab:results_compare}. We don't directly compare to other related work since their performance is reported with access to gold-entities and their quality when building a practical QA system with an external ER is unknown.

Compared to \textit{Rigel-Baseline}, there is approximately a $3\%$ drop in performance when gold question entities are not provided to the model (\textit{Rigel-ER}). We realized this is mainly due to cases where it is not possible to distinguish between all possible candidate entities based on the question alone. This is consistent with earlier studies that conclude $15 \text{-}17\%$ of questions in these datasets cannot be answered due to context ambiguity \cite{han-etal-2020-empirical, petrochuk-zettlemoyer-2018-simplequestions}. For example, in the question “What position does \textit{carlos gomez} play?” ("carlos gomez" given as correct span), Rigel-ER learns to give higher likelihood to athletes compared to art performers; but since the question does not include discriminative information such as sport or team name, all athletes called "Carlos Gomez" will receive very similar likelihood scores. 

There is a further drop in performance when we go from \textit{Rigel-ER }to \textit{Rigel-E2E}, which performs full E2E learning. This time, the errors can be explained by the fact that different spans produce candidates with overlapping entity types, leaving the model with little signal to prefer one span over another. 

For example, given the question “who directed the film gone with the wind?”, \textit{Rigel-ER} is given the correct span “gone with the wind” and just needs to disambiguate between \code{Q2875} (the Wikidata entity ID for the American film "Gone with the Wind") and the other candidates stemming from that span. \textit{Rigel-E2E} will additionally need to learn to maximize the span score (Equation \ref{eq:md2}) for “gone with the wind” compared to other spans in the question, such as “the film”, “the wind”, and “wind”, which are all film titles as well. This is a difficult task since all these spans produce film entities, and relying on the loss from following the \textit{director} relation is not enough to effectively disambiguate between them.


We are working on a few solutions to alleviate this span ambiguity issue with \textit{Rigel-E2E}. The main question is, what should we do when span scores are diffused and not spiked? This, for example, happens in the above question and the 4 spans: “gone with the wind” , “the film”, “the wind”, and “wind”. A simple post-processing step to merge overlapping spans seems to be quite effective. In the example above, "the wind" and "wind" fall under "gone with the wind", and given that their scores are similar we can decide to assign all child span scores to their parent. Diversity or entropy of candidates produced by a certain span also seems to be helpful in pruning bad spans. In the above question, candidate entities from the span “wind” include movies, companies, music bands, and even a satellite, among others. On the other hand, candidate entities for “gone with the wind” are mostly works of art, suggesting that it may be a better choice. We are looking into using this information as part of training, as well as post-processing.

While we don't directly compare, the gap between our results and other related work is partly due to the inference mechanism used. At this time, ReifiedKB only supports a relation following operation (Equation \ref{eq:reified-follow-kgqa}), while, for instance, EmQL \cite{sun2020faithful}  additionally supports set intersection, union, and difference. These additional operations allow answering more complex questions present in the WebQSP dataset. We are working on adding support for intersection, union, count, min, and max operations to our model as future work.

We’d like to emphasize that although including the ER component adversely affects the results, extracting question entities is a necessity for real world applications, and alternatives with off-the-shelf models do perform worse. Hence, we believe our approach is more practical, especially given the lack of training data for ER on questions.








\begin{table}[t]
\begin{center}
\begin{tabular}{l c c}
\toprule
Model & WQSP & SIQ  \\
\midrule
KVMem \cite{miller-etal-2016-key} & 46.7 & - \\
ReifiedKB \cite{Cohen2020ICLR} & 52.7 & - \\
EmQL \cite{sun2020faithful} & 75.5 & - \\
MemNN \citep{DBLP:journals/corr/BordesUCW15} & - & 61.6 \\
KBQA-Adapter  \cite{wu-etal-2019-learning} & - & 72.0  \\


\midrule
Rigel-Baseline & 52.4 & 73.4 \\
Rigel-ER   & 48.2 & 70.1 \\
Rigel-E2E  & 45.0 & 68.2 \\
\bottomrule
\end{tabular}
\end{center}
\caption{Comparison of Hits@1 results on WebQSP (WQSP) and Accuracy on SimpleQuestions (SIQ)}
\label{tab:results_compare}
\end{table}

\section{Conclusion}

In this work, we proposed a solution for KGQA that jointly learns to perform entity resolution (ER) and multi-hop inference. Our model extends the boundaries for end-to-end learning and is weakly supervised using pairs of only questions and answers.  This eliminates the need for external components and expensive domain-specific labelled data for ER. We further demonstrate the feasibility of this approach on two open-domain QA datasets. 

\bibliography{anthology,custom}
\bibliographystyle{acl_natbib}

\appendix
\section{Model Hyperparameters}

We train Rigel models using the hyperparameters below on a single GPU machine with 16GB GPU memory (AWS p3.2xlarge). WebQSP requires more than 1 hop for question answering, leading to a larger knowledge graph, so we use a smaller batch size to avoid out of memory issues. Training with early stopping completes in approximately 4-7 hours depending on the model configuration used (Rigel-baseline, Rigel-ER, Rigel-E2E).

\begin{table}[hbt!]
    \begin{center}
        \begin{tabular}{l r r}
            \toprule
            Hyperparameter & SIQ & WQSP \\
            \midrule
            Batch Size & 32 & 6\\
            Gradient Accumulation & 8 & 32\\
            Max Training Steps & 20000 & 30000\\
            Learning Rate & 1e-4 & 1e-4\\
            Max Number of Hops & 1 & 3\\
            \bottomrule
        \end{tabular}
    \end{center}
    \caption{Hyperparameters for training on WebQuestionsSP (WQSP) and SimpleQuestions (SIQ)}
    \label{tab:hyperparam}
\end{table}

\section{Examples}

The table below shows outputs of Rigel-E2E model on two questions from SimpleQuestions. In the first example, the model assigns high likelihood to the correct span and candidate entity. The inference module also assigns a high likelihood to the right relation (\texttt{instance of}), which leads to the correct answer entity.

In the second question, the model assigns higher likelihood to the \texttt{sam edwards} span, but it's not very confident and other spans such as \texttt{sam} and \texttt{edwards} receive similar scores. In addition, there's a large overlap between candidate entities of these spans (i.e. all produce candidates which are human and have \texttt{place of birth} property). This ambiguity in context leads to the ground truth question entity receiving a low likelihood. Even though the right relation is predicted by the inference module, the final answer entity is different from the answer label.

\begin{table*}[b]
\begin{center}
\begin{tabular}{l}
\toprule
\textbf{Question:} what is the category of the celestial object 1241 dysona? \\
\textbf{Question Entity:} Q137259 (1241 Dysona) \\ 
\textbf{Answer Entity} Q3863 (asteroid) \\  \\

Span Likelihoods: \\
\textbf{ ('1241 dysona', 0.965)}, ('celestial', 0.013), ('celestial object', 0.011), \\
('object', 0.009), ('1241', 0.002), ('the category', 0.0), ('what is', 0.0), ('category', 0.0) \\ \\

Candidate Entity Likelihoods: \\
\textbf{('Q137259', 0.998)}, ('Q6999', 0.0), ('Q66311333', 0.0), ('Q488383', 0.0), $\cdots$  \\ \\

Top Prediction:\\
('1241 dysona', 0.965) $\rightarrow$ \textbf{(instance of, 1.000)  } $\rightarrow$ ('Q3863', 0.998) \cmark  \\

\midrule

\textbf{Question:} what is the place of birth of sam edwards? \\
\textbf{Question Entity:} Q472382 (Sam Edwards, Welsh Physicist) \\ 
\textbf{Answer Entity} Q23051 (Swansea) \\  \\

Span Likelihoods: \\
('edwards', 0.366), \textbf{('sam edwards', 0.332)}, ('sam', 0.301), ('what is', 0.0), ('the place', 0.0), \\ 
('place', 0.0), ('the place of birth', 0.0), ('place of birth', 0.0), ('birth', 0.0) \\ \\

Candidate Entity Likelihoods: \\ 
('Q3470479', 0.25), ('Q835638', 0.111), ('Q911493', 0.058), ('Q2691159', 0.017), \\
('Q20812281', 0.016), ('Q1816301', 0.014), ('Q47465190', 0.013), ('Q1118055', 0.011), \\
('Q58317511', 0.01), \textbf{('Q472382', 0.01)}, ('Q27925002', 0.01), ('Q52852726', 0.009), $\cdots$ \\ \\

Top Prediction: \\
('Q3470479', 0.25) $\rightarrow$ \textbf{(place of birth, 1.000)} $\rightarrow$ ('Q219656', 0.250) \xmark  \\

\bottomrule
\end{tabular}
\end{center}
\caption{Example outputs of Rigel-E2E on SimpleQuestions}
\label{tab:examples}
\end{table*}

\end{document}

%% file: math_commands.tex
\usepackage{amsmath,amsfonts,bm}

\newcommand{\code}[1]{\texttt{#1}}

\def\eqref#1{equation~\ref{#1}}

\def\1{\bm{1}}

\DeclareMathAlphabet{\mathsfit}{\encodingdefault}{\sfdefault}{m}{sl}
\SetMathAlphabet{\mathsfit}{bold}{\encodingdefault}{\sfdefault}{bx}{n}

